 \documentclass[authoryear,preprint,review,12pt]{elsarticle}
\usepackage{amsmath,amssymb, balance, subfigure, makecell, array, url, algorithmic}
\newcommand{\argmax}{\operatornamewithlimits{argmax}} 
\newcolumntype{x}[1]{
>{\centering\hspace{0pt}}p{#1}}
\newcolumntype{C}[1]{>{\centering\let\newline\\\arraybackslash\hspace{0pt}}m{#1}}
\usepackage[utf8]{inputenc}
\usepackage{algorithm}
\usepackage[round]{natbib}
\journal{Knowledge-Based Systems}

\begin{document}

\begin{frontmatter}

\title{An ensemble-based system for automatic screening of diabetic retinopathy}


\author{B\'alint Antal, Andr\'as Hajdu}

\address{University of Debrecen,	Faculty of Informatics\\ 4010 Debrecen, POB 12, Hungary.\\
Email: {\tt\small \{antal.balint, hajdu.andras\}@inf.unideb.hu}}

\begin{abstract}
In this paper, an ensemble-based method for the screening of diabetic retinopathy (DR) is proposed. This approach is based on features extracted from the output of several retinal image processing algorithms, such as image-level (quality assessment, pre-screening, AM/FM), lesion-specific (microaneurysms, exudates) and anatomical  (macula, optic disc) components. The actual decision about the presence of the disease is then made by an ensemble of machine learning classifiers. We have tested our approach on the publicly available Messidor database, where 90\% sensitivity, 91\% specificity and 90\% accuracy and 0.989 AUC are achieved in a disease/no-disease setting. These results are highly competitive in this field and suggest that retinal image processing is a valid approach for automatic DR screening.

\end{abstract}


\end{frontmatter}

\section{Introduction}
\label{sec:introduction}

Diabetic retinopathy (DR) is a consequence of diabetes mellitus which manifests itself in the retina. This disease is one of the most frequent causes of visual impairment in developed countries and is the leading cause of new cases of blindness in the working age population. In 2011, 366 million people were diagnosed with diabetes and a further 280 million people were having risk to develop it. At any point in time, approximately 40\% of diabetic patients suffer from DR, out of which an estimated 5\% face the sight-threatening form of this disease. Altogether, nearly 75 people go blind every day as a consequence of DR even though treatment is available.   

Automatic computer-aided screening of DR is a highly investigated field \Citep{system}. The motivation for creating reliable automatic DR screening systems is to reduce the manual effort of mass screening \Citep{Fleming2011}, which also raises a financial issue \Citep{Scotland2010}. While several studies focus on the recognition of patients having DR \Citep{Fleming2011} \Citep{early} and considering the specificity of the screening as a matter of efficiency, we show how both sensitivity and specificity can be kept at high level by combining novel screening features and a decision-making process. Especially, our results are very close to meet the recommendations of the British Diabetic Association (BDA) (80\% sensitivity and 95\% specificity \Citep{bda}). 

The basis for an automatic screening system is the analysis of color fundus images \Citep{abramoff_review}. The key to the early 
recognition of DR is the reliable detection of microaneurysms (MAs) on the retina, which serves as an essential part for most automatic DR screening systems \Citep{early} \Citep{Jelinek2006} \Citep{antal_tbme2012} \Citep{Niemeijer2009}. The role of bright lesions for DR grading has also been investigated with positive \Citep{Fleming2010} and negative outcomes \Citep{early} reported. Besides lesions, image quality assesment \Citep{Philip} \Citep{largescale} is also considered to exclude ungradeable images. As a new direction, in \Citep{Agurto} an image-level DR recognition algorithm is also presented. 

The proposed framework extends the state-of-the-art components of an automatic DR screening system by adding pre-screening \Citep{antal_jocs} and the distance of the macula center (MC) and the optic disc center (ODC) as novel components. We also use image quality assessment as a feature for classification rather than a tool for excluding images. The comparison of the components used in some recently published automatic DR screening systems can be found in Table \ref{tab:components}.   

\begin{table*}[!ht]
\caption{Comparison of components of the automatic screening system.}
\centering
\begin{tabular}{|c|C{0.1\linewidth}|C{0.1\linewidth}|C{0.1\linewidth}|C{0.1\linewidth}|C{0.1\linewidth}|C{0.1\linewidth}|}
\hline
Screening system & Image quality & Red lesion & Bright lesion & AM/FM & Pre-screening & MC-ODC\\ 
\hline
\Citep{early} &  & X &  &  & &\\
\hline
\Citep{Jelinek2006} & & X & & & &\\
\hline
\Citep{antal_tbme2012} &  & X &  &  & &\\
\hline
\Citep{Philip} & X & X & & & &\\
\hline
\Citep{largescale} & X & X & X & & &\\
\hline
\Citep{Agurto} &  &  &  & X & &\\
\hline
Proposed & X & X & X & X & X & X\\
\hline
\end{tabular}
\label{tab:components}
\end{table*}

Regarding decision making, automatic DR screening systems either partially follow clinical protocols (e.g. MAs indicate presence of DR) \Citep{Jelinek2006} \Citep{antal_tbme2012} \Citep{Philip} \Citep{largescale} or use a machine learning classifier \Citep{system} \Citep{Fleming2010} \Citep{Agurto}. A common way to improve reliability in machine learning based applications is to use ensemble-based approaches \Citep{kuncheva}. For medical decision support, ensemble methods have been successfully applied to several fields. 
In \Citep{West2005} the authors have investigatedthe applicability  of ensembles for breast cancer data classification. The prediction of response to certain therapy is improved by the use of a classifier ensemble \Citep{Moon2007}.
In \Citep{Eom2008} the authors used an ensemble of four classifiers for cardiovascular disease prediction.
Ensemble methods are also provided improvement over single classifiers in a natural language processing environment \Citep{Doan2012}.

Ensemble systems combine the output of multiple learners with a specific fusion strategy. In \Citep{early} and \Citep{antal_tbme2012}, the fusion of multiple MA detectors has proven to be more efficient than a single algorithm for DR classification. The proposed system is ensemble-based at more levels: we consider ensemble systems both in image processing tasks and decision making.  

\begin{figure*}[htb]
	\centering
		\includegraphics[width=\linewidth]{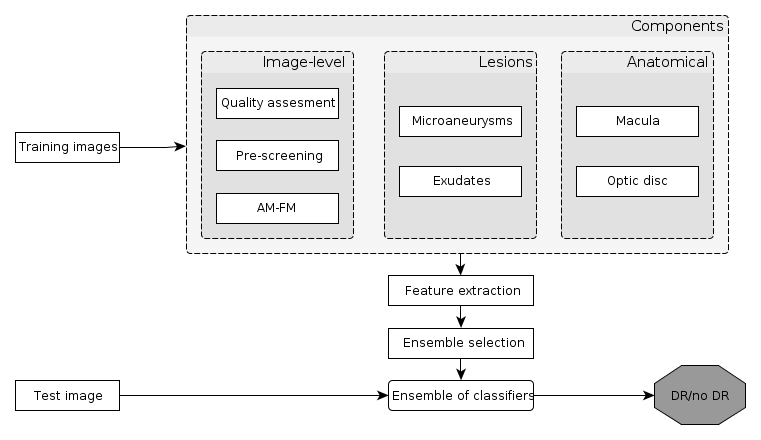}
	\caption{Flow chart of the proposed decision support framework.}
	\label{fig:final_detection_flow}
\end{figure*}

In this paper, a framework for the automatic grading of color fundus images regarding DR is proposed. The approach classifies images based on characteristic features extracted by lesion detection and anatomical part recognition algorithms. These features are then classified using an ensemble of classifiers. As the results show, the proposed approach is highly efficient for this task. The flow chart of our decision making
protocol can be seen in Figure \ref{fig:final_detection_flow}, as well.

We have tested our approach on the publicly available dataset Messidor (see http://messidor.crihan.fr), where it has provided a 0.989 area under the ROC curve (AUC) value in a disease/no disease setting, which is a relatively high figure compared with other state-of-the-art techniques.

The rest of the paper is organized as follows: in section \ref{sec:components}, we present the image processing components of our system. Section \ref{sec:ensemble} presents the details of the presented ensemble learning framework. Our experimental methodology and results can be found in sections \ref{sec:methodology} and \ref{sec:exp_final}, respectively. Finally, we draw conclusions in section \ref{sec:conclusion}.  

\section{Components of an automatic system for diabetic retinopathy screening}
\label{sec:components}

In this section, the components we used for feature extraction are described. They can be classified as image-level, lesion-specific, and
anatomical ones.

\subsection{Image-level components}
\label{sec:imagelevel}

\subsubsection{Quality assessment}
\label{sec:qa}

We classify the images whether they have sufficient quality for a reliable decision with a supervised classifier, where the box count values of the detected vessel system serve as features  \Citep{antal_wisp2009}. For vessel segmentation we use an approach proposed in \Citep{gyuri} based on Hidden Markov Random Fields (HMRF). Here, the authors extend the optimization problem of HMRF models considering the tangent vector field of the image
to enhance the connectivity of the vascular system consisting of elongated structures. 

\subsubsection{Pre-screening}
\label{sec:prefilter}

During pre-screening \Citep{antal_jocs}, we classify the images as severely diseased (abnormal) ones or to be forwarded for further processing. Each image is split into disjoint regions and a simple texture descriptor (inhomogeneity measure) is extracted for each region. Then, a machine learning classifier is trained to classify the images based on these features.

\subsubsection{Multi-scale AM/FM based feature extraction}
\label{sec:amfm}

The Amplitude-Modulation Frequency-Modulation (AM/FM)  \Citep{amfm} method extracts
information from an image, decomposing the
green channels of the images into different representations
which reflect the intensity, geometry, and texture of the structures
with signal processing techniques. The extracted information are then filtered to establish 39 different representations of the image. The images are classified using these features with a supervised learning method. More on this approach can be found in \Citep{amfm}. 

\subsection{Lesion-specific components}
\label{sec:lesionspecific}
 
\subsubsection{Microaneurysm detection}
\label{sec:ma}

Microaneurysms are normally the earliest signs of DR. They appear as small red dots in the image and their resemblance to vessel fragments make it hard  to detect them efficiently. In the proposed system, we apply the MA detection method described in \Citep{antal_tbme2012}, which is an efficient approach based on $\langle$preprocessing method, candidate extractor$\rangle$ ensembles. 
 
\subsubsection{Exudate detection}
\label{sec:exudate}

Exudates are primary signs of diabetic retinopathy and occur when lipid or fat leak from blood vessels or aneurysms. Exudates are bright, small spots, which can have irregular shape. Since exudate detection is also a challenging task, we follow the same complex methodology as for MA detection \Citep{antal_pr2012}. Thus, we combine preprocessing methods and candidate extractors in the case of exudate detection, as well \Citep{antal_ispa2011}.  

\begin{figure}[htb]
\centering
\subfigure[Microaneurysm]{
\label{fig:ma}
\includegraphics[keepaspectratio,width=0.25\linewidth]{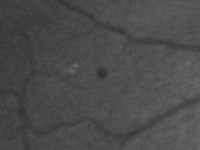}
}
\quad
\subfigure[Exudates]{
\label{fig:ex}
\includegraphics[keepaspectratio,width=0.25\linewidth]{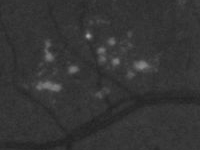}
}
\quad
\subfigure[Inhomogeneity]{
\label{fig:inh}
\includegraphics[keepaspectratio,width=0.25\linewidth]{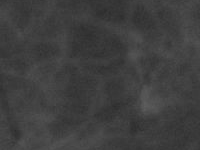}
}
\caption{Some representative visual features to be extracted from the images.}
\label{fig:features}
\end{figure}

In Figure \ref{fig:features}, we show some examples for the appearance of DR-related symptoms in retinal images.

\subsection{Anatomical components}
\label{sec:anatomical}

\subsubsection{Macula detection}
\label{sec:macula}

The macula is the central region of sharp vision in the human eye with its center referred to as the fovea. Any lesions appearing within the macula can lead to severe loss of vision. Therefore, the efficient detection of the macula is essential in an automatic screening system for DR. The macula is located roughly in the center of the retina, temporal to the optic nerve. In our system, we use the method described in  \Citep{antal_acta2011}, which extracts the largest component from the image which is darker than its surroundings.
The location of the macula together with the optic disc described below define some features incorporated in our decision framework.
  
\subsubsection{Optic disc detection}
\label{sec:od}

The optic disc is a circular shaped anatomical structure with a bright appearance. It is the area, where the optic
nerve enters the eye. If the center and the radius of the optic disc are detected correctly, they can be used as
reference data for locating other anatomical parts e.g. the macula. In our system, we use the ensemble-based system described in \Citep{Qureshi2011}.
Recognizing these anatomical parts is important from two aspects: the appearance of certain lesions at specific positions can indicate a more advanced stage of DR and the presence of rare, but serious defects (like retinal detachment) can ruin the detection of the optic disc and macula.  

\section{Ensemble learning}
\label{sec:ensemble}

The most important expectation for a computer-aided medical system is its high reliability. To ensure that, we use ensemble-based decision making \Citep{kuncheva}. Thus, we have trained several classifiers to separate DR and non-DR cases and fused their results. In this section, we describe how we select the ensemble for DR classification based on the features extracted from the output of the detectors presented in section \ref{sec:components}.

\subsection{Concepts of ensemble learning}
\label{sec:ens_learning_concepts}

The basic concepts of ensemble learning are presented by following the classic literature \Citep{kuncheva}. 
These concepts formalize our ensemble-based system for DR grading described in the forthcoming sections.

\newcounter{qc}
\newtheorem{classifier}[qc]{Definition}
\begin{classifier}
\label{classifierdef}
Let $\Omega = \left\{ \omega_{1}, \omega_{2}, \dots, \omega_{M}\right\}$ be a set of class labels. Then, a function $D: \mathbb{R}^{n} \rightarrow \Omega$ 
is called a classifier, while a vector $\vec{\chi} = \left(\chi_{1}, \chi_{2}, \dots, \chi_{n}\right) \in \mathbb{R}^{n}$ is called a feature vector.
\end{classifier}

\newtheorem{discrimination}[qc]{Definition}
\begin{discrimination}
Let $h_{1}, h_{2}, \dots, h_{M}$, $h_{i}: \mathbb{R}^{n} \rightarrow \mathbb{R}$, $i = 1,\,\dots,\,M$ be so-called discriminator functions corresponding to the class labels $\omega_{1}, \omega_{2}, \dots, \omega_{M}$, respectively. Then, the classifier $D$ belonging to these
discriminator functions is defined by:
\begin{equation}
	D\left(\vec{\chi}\right) = \omega_{j*} \iff h_{j*}\left(\vec{\chi}\right) = \max_{j = 1}^{M}\left(h_{j}\left(\vec{\chi}\right)\right).
\end{equation}
for all $\vec{\chi}\in \mathbb{R}^{n}$.
\end{discrimination}

\newtheorem{voting}[qc]{Definition}
\begin{voting}
\label{def:ec_maj}
Let $D_{1}, D_{2}, \dots, D_{L}$ be classifiers. Then, the majority voting ensemble classifier ${\cal D}_{maj} : \mathbb{R}^{n} \rightarrow \Omega$ formed from these classifiers is defined as:
\begin{equation}
{\cal D}_{maj} \left(\vec{\chi}\right) = \omega_{{i^*}} \iff 
\left|\left\{j: D_{j}\left(\vec{\chi}\right) = \omega_{{i^*}}, j=1,\dots,M \right\}\right| =
\max_{i = 1}^{M}\left|\left\{j: D_{j}\left(\vec{\chi}\right) = \omega_{i}, j=1,\dots,M \right\}\right|.
\end{equation}
\end{voting}

\newtheorem{weighted_voting}[qc]{Definition}
\begin{weighted_voting}
\label{def:ec_wmaj}
Let $D_{1}, D_{2}, \dots, D_{L}$ be classifiers and $\vec{\beta} = \left(\beta_{1}, \beta_{2}, \dots, \beta_{L}\right) \in \mathbb{R}^L$ be a weight vector assigned to the classifiers. Then, the weighted majority voting ensemble classifier ${\cal D}_{wmaj} : \mathbb{R}^{n} \rightarrow \Omega$ is defined as follows:
\begin{equation}
{\cal D}_{wmaj} \left(\vec{\chi}\right) = \omega_{{i^*}} \iff 
\sum_{\substack{j=1\\ D_{j}\left(\vec{\chi}\right) = \omega_{{i^*}}}} ^L \beta_j=
\max_{i = 1}^{M}\left( \sum_{\substack{j=1\\ D_{j}\left(\vec{\chi}\right) = \omega_{i}}} ^L \beta_j \right).
\end{equation}
\end{weighted_voting}

\newtheorem{arithmetic_voting}[qc]{Definition}
\begin{arithmetic_voting}
\label{def:ec_arv}
Let $D_{1}, D_{2}, \dots, D_{L}$ be classifiers and $h_{j,i}$ be a discriminator function of the classifier $D_{j}$ for the class $i$, $i = 1,\dots,M,\,j = 1,\dots,L$. Then, the following algebraic ensemble classifiers can be defined:
\begin{align}
{\cal D}_{avg} \left(\vec{\chi}\right) &= \omega_{{i^*}} \iff \dfrac{1}{L} \sum_{j = 1}^{L}\left(h_{j,{i^*}}\left(\vec{\chi}\right)\right) =
\max_{i = 1}^{M}\left(\dfrac{1}{L} \sum_{j = 1}^{L}\left(h_{j,i}\left(\vec{\chi}\right)\right)\right),\\
{\cal D}_{pro} \left(\vec{\chi}\right) &= \omega_{{i^*}} \iff \prod_{j = 1}^{L}\left(h_{j,{i^*}}\left(\vec{\chi}\right)\right) =
\max_{i = 1}^{M}\left(\prod_{j = 1}^{L}\left(h_{j,i}\left(\vec{\chi}\right)\right)\right),\\
{\cal D}_{min} \left(\vec{\chi}\right) &= \omega_{{i^*}} \iff \min_{j = 1}^{L}\left(h_{j,{i^*}}\left(\vec{\chi}\right)\right) =
\max_{i = 1}^{M}\left(\min_{j = 1}^{L}\left(h_{j,i}\left(\vec{\chi}\right)\right)\right),\\
{\cal D}_{max} \left(\vec{\chi}\right) &= \omega_{{i^*}} \iff \max_{j = 1}^{L}\left(h_{j,{i^*}}\left(\vec{\chi}\right)\right) = 
\max_{i = 1}^{M}\left(\max_{j = 1}^{L}\left(h_{j,i}\left(\vec{\chi}\right)\right)\right).
\end{align}
\end{arithmetic_voting} 

\subsection{Ensemble selection}
\label{sec:selection}

To select the optimal ensemble for DR classification, we have trained several well-known classifiers that will be
described in section \ref{sec:classifiers}. Each ensemble is a subset of these classifiers. Several approaches have been tested for selecting the best subset of classifiers ${\cal D}$ for DR grading. The following search methods were investigated based on \Citep{Ruta} for a fixed set of classifiers $\{D_{1},\,\dots,\,D_{L}\}$ and energy function $E : {\cal D} \subseteq \{D_{1},\,\dots,\,D_{L}\} \to {\mathbb R}_{\geq 0}$: 
\begin{itemize}
	\item \textbf{Forward search:} 	First, the best individual classifier  is selected. Then, further classifiers are added if the performance of the ensemble increases. The process ends when no further increase of performance is reached by adding more classifiers. Algorithm \ref{alg:forward}
gives a formal description of this search method.
	\begin{algorithm}
	\caption{Forward search}
	 \label{alg:forward} 
	\begin{algorithmic}[1]
\algsetup{linenodelimiter=.}
		\STATE ${\cal D} \leftarrow \argmax\left(E\left(\{D_{1}\}\right), E\left(\{D_{2}\}\right), \dots, E\left(\{D_{L}\}\right)\right)$	\\
		\STATE $e_{best} \leftarrow E\left({\cal D}\right)$\\
		\FORALL{$D_{i} \notin {\cal D}, i = 1\,\dots\,L$}			
			\STATE $e \leftarrow E\left({\cal D} \cup \{D_{i}\}\right)$\\
			\IF {$e > e_{best}$}
					\STATE ${\cal D} \leftarrow  {\cal D} \cup  \{D_{i}\}$\\
					\STATE $e_{best} \leftarrow e$\\
			\ENDIF		
		\ENDFOR
		\RETURN ${\cal D}$
	\end{algorithmic}	
	\end{algorithm}

	\item \textbf{Backward search:} First, all classifiers are considered as members of the ensemble. Then, classifiers are removed from the ensemble while the performance of the ensemble increases. See Algorithm \ref{alg:backward} for a formal description.
		\begin{algorithm}
	\caption{Backward search}
	 \label{alg:backward} 
	\begin{algorithmic}[1]
\algsetup{linenodelimiter=.}
		\STATE ${\cal D} \leftarrow \{D_{1},\,\dots,\,D_{L}\}$	\\
		\STATE $e_{best} \leftarrow E\left({\cal D}\right)$\\
		\FORALL{$D_{i} \in {\cal D}$}			
			\STATE $e \leftarrow E\left({\cal D} \setminus \{D_{i}\}\right)$\\
			\IF {$e > e_{best}$}
					\STATE ${\cal D} \leftarrow  {\cal D} \setminus  \{D_{i}\}$\\
					\STATE $e_{best} \leftarrow e$\\
			\ENDIF		
		\ENDFOR
		\RETURN ${\cal D}$
	\end{algorithmic}	
	\end{algorithm}
\end{itemize}	

For comparison, we also consider the following two ensembles besides the ones found by the search methods:
\begin{itemize}
	\item \textbf{All:} All classifiers are members of the ensemble.
	\item \textbf{Single best:} The ensemble contains only the best performing classifier.
\end{itemize}

\section{Methodology}
\label{sec:methodology}

\subsection{Messidor database}
\label{sec:messidor_db}

\begin{figure*}[htb]
\centering
\subfigure[R0]{
\label{fig:r0}
\includegraphics[keepaspectratio,width=0.3\linewidth]{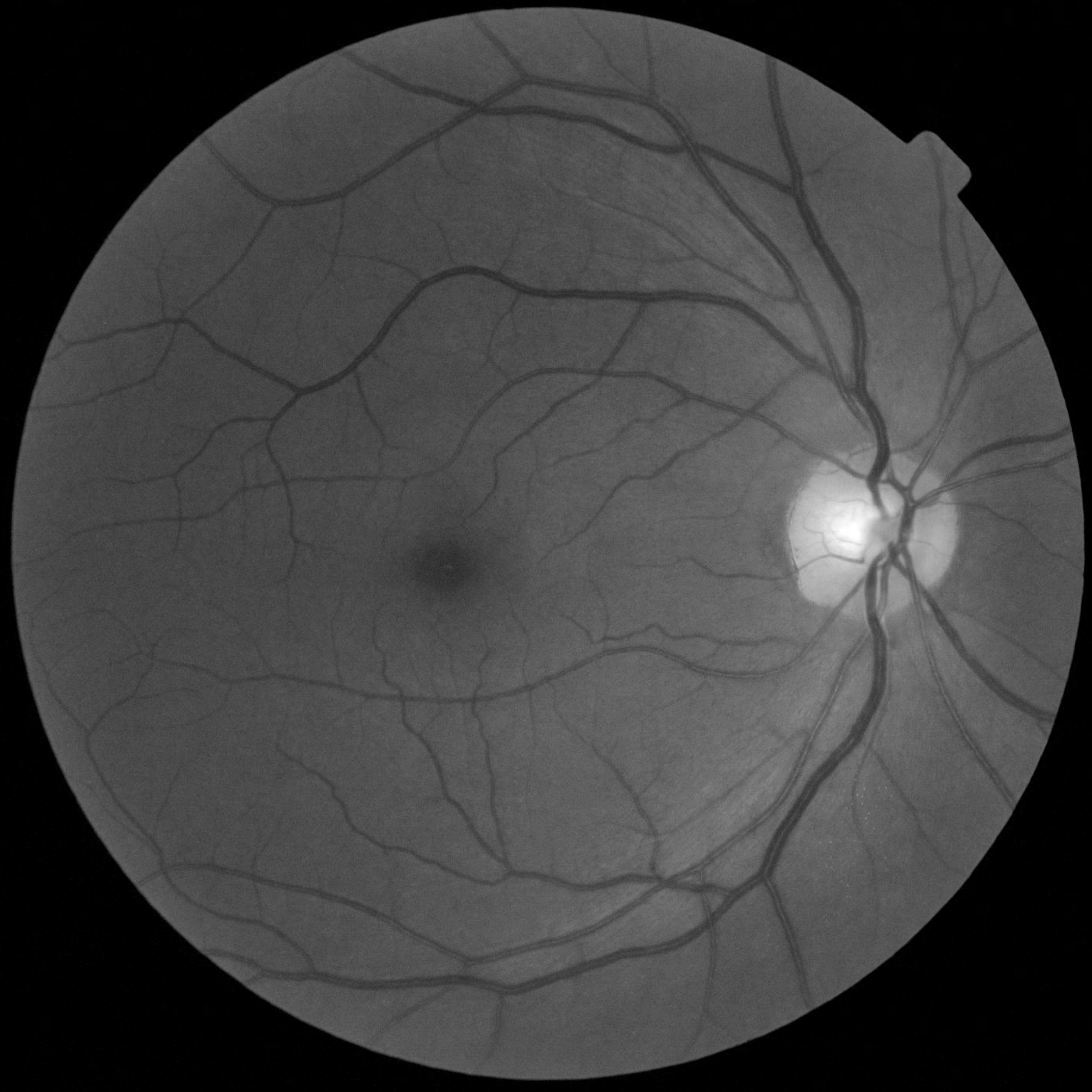}
}
\quad
\subfigure[R1]{
\label{fig:r1}
\includegraphics[keepaspectratio,width=0.3\linewidth]{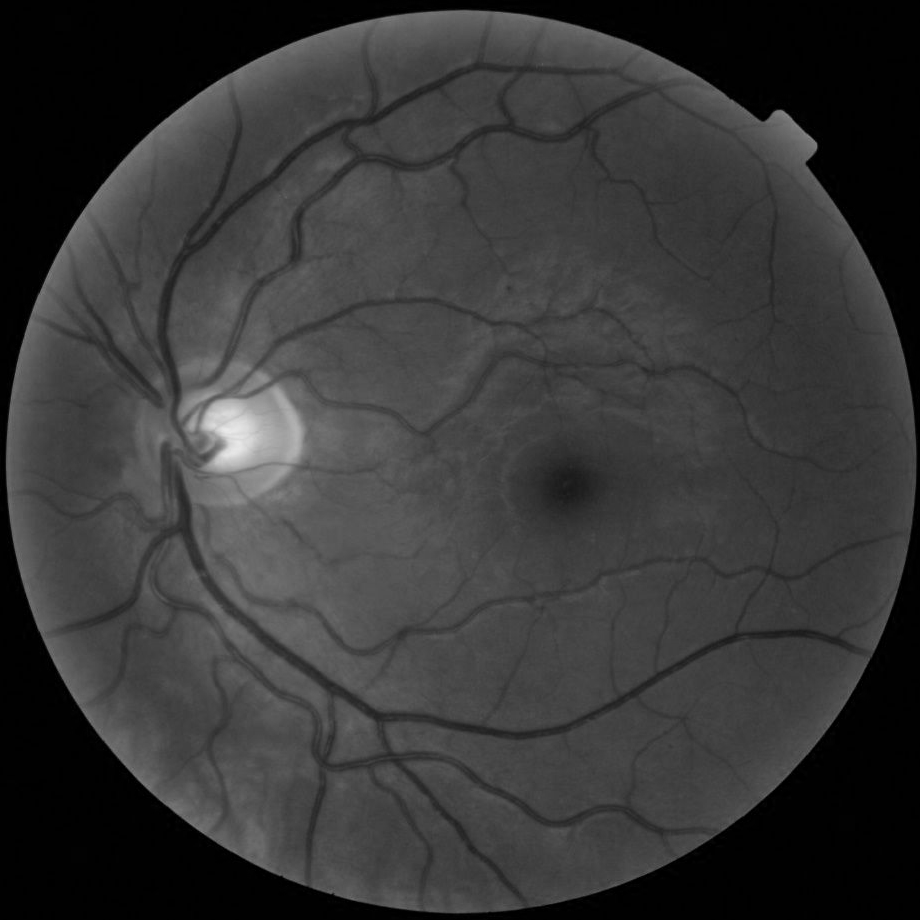}
}
\\
\subfigure[R2]{
\label{fig:r2}
\includegraphics[keepaspectratio,width=0.3\linewidth]{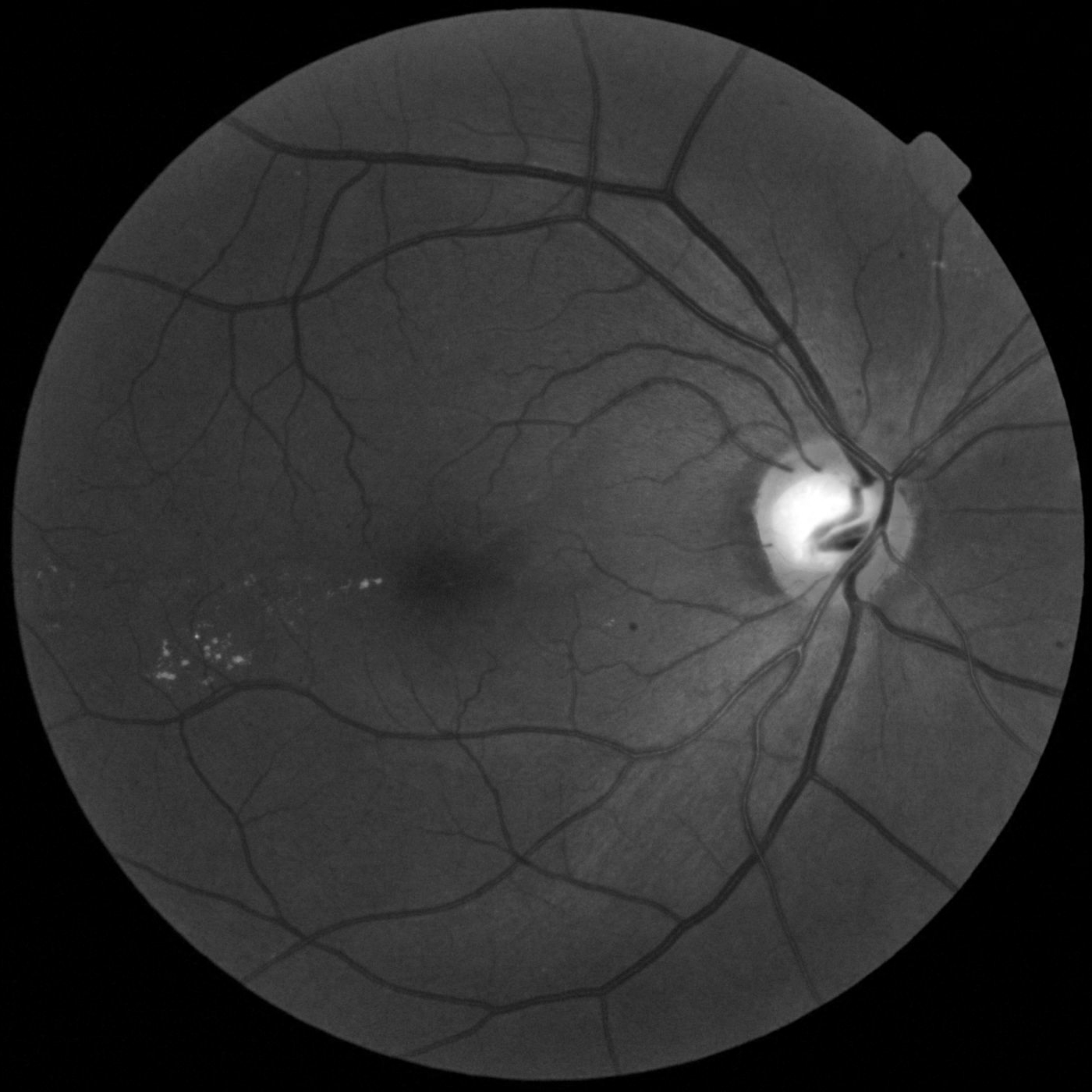}
}
\quad
\subfigure[R3]{
\label{fig:r3}
\includegraphics[keepaspectratio,width=0.3\linewidth]{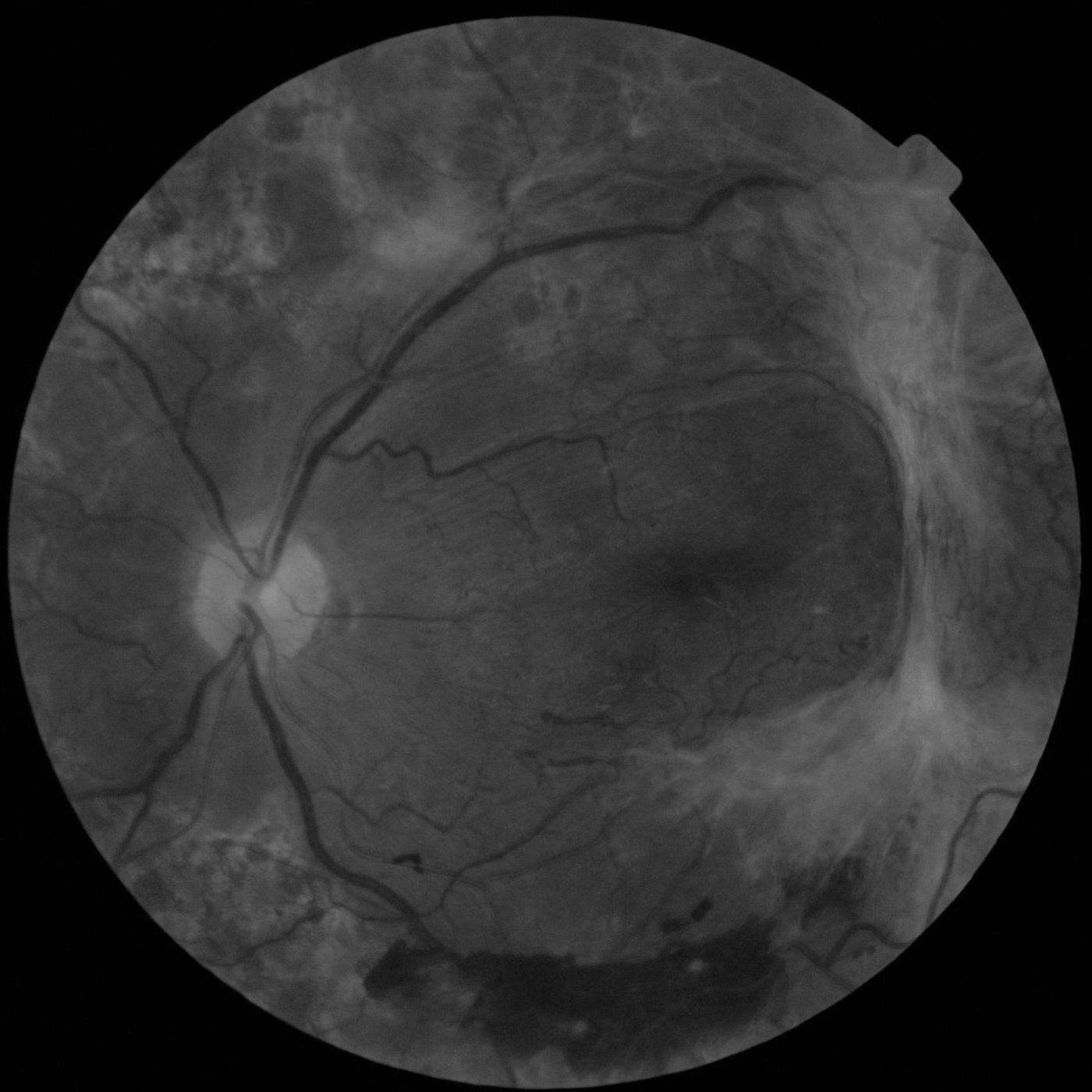}
}
\caption{Representative images having different grades (R0, R1, R2, R3) from the Messidor database.}
\label{fig:grades}
\end{figure*}

For experimental studies, we consider the publicly available Messidor database that consists of 1200 losslessly compressed images with $45^{\circ}$ FOV and different resolutions ($440 \times 960$, $2240 \times 1488$ and $2304 \times 1536$ pixels). For each image, a grading score ranging from R0 to R3 is also provided. These grades correspond to the following clinical conditions: a patient with grade R0 has no DR. R1 and R2 are mild and severe cases of non-proliferative retinopathy, respectively. Finally, R3 stands for the most serious condition. The grading is based on the appearance of MAs, haemorrhages and neovascularization. The corresponding proportion of the images in the Messidor dataset: 540 R0 (46\%), 153 R1 (12.75\%), 247 R2 (20.58\%) and 260 R3 (21.67\%). This database is made available by the Messidor program partners (see http://messidor.crihan.fr).
Some example images corresponding to the different grading scores are shown in Figure \ref{fig:grades}.

\subsection{Features}
\label{sec:features}

In this section, we describe the features that were extracted from the output of the image processing algorithms presented in section \ref{sec:components}. These features are also summarized in Table \ref{tab:dr_features}.

$\bullet ~\chi_{0}$ is the result of quality assessment, which is a real number between 0 (worst) and 1 (best) quality for a color fundus image. 

$\bullet ~\chi_{1}$ is a binary variable representing the result of pre-screening, where 1 indicates severe retinal abnormality and 0 its lack. 

$\bullet ~$ As an essential part of a DR screening system, features $\chi_{2}-\chi_{7}$ describe the results of MA detection. 
More precisely, $\chi_{i}$ ($i=2,\dots,7$) stand for the number of MAs found at the confidence levels $\alpha=0.5,\dots,1$, respectively.

$\bullet ~\chi_{8} - \chi_{16}$ contain the same information as $\chi_{2}-\chi_{7}$ for exudates. However, as exudates are represented by a set of points rather than the number of pixels constructing the lesions, these features are normalized by dividing the number of lesions with the diameter of the ROI to compensate different image sizes.

\begin{figure}[!ht]
	\centering
		\includegraphics[width=0.6\linewidth]{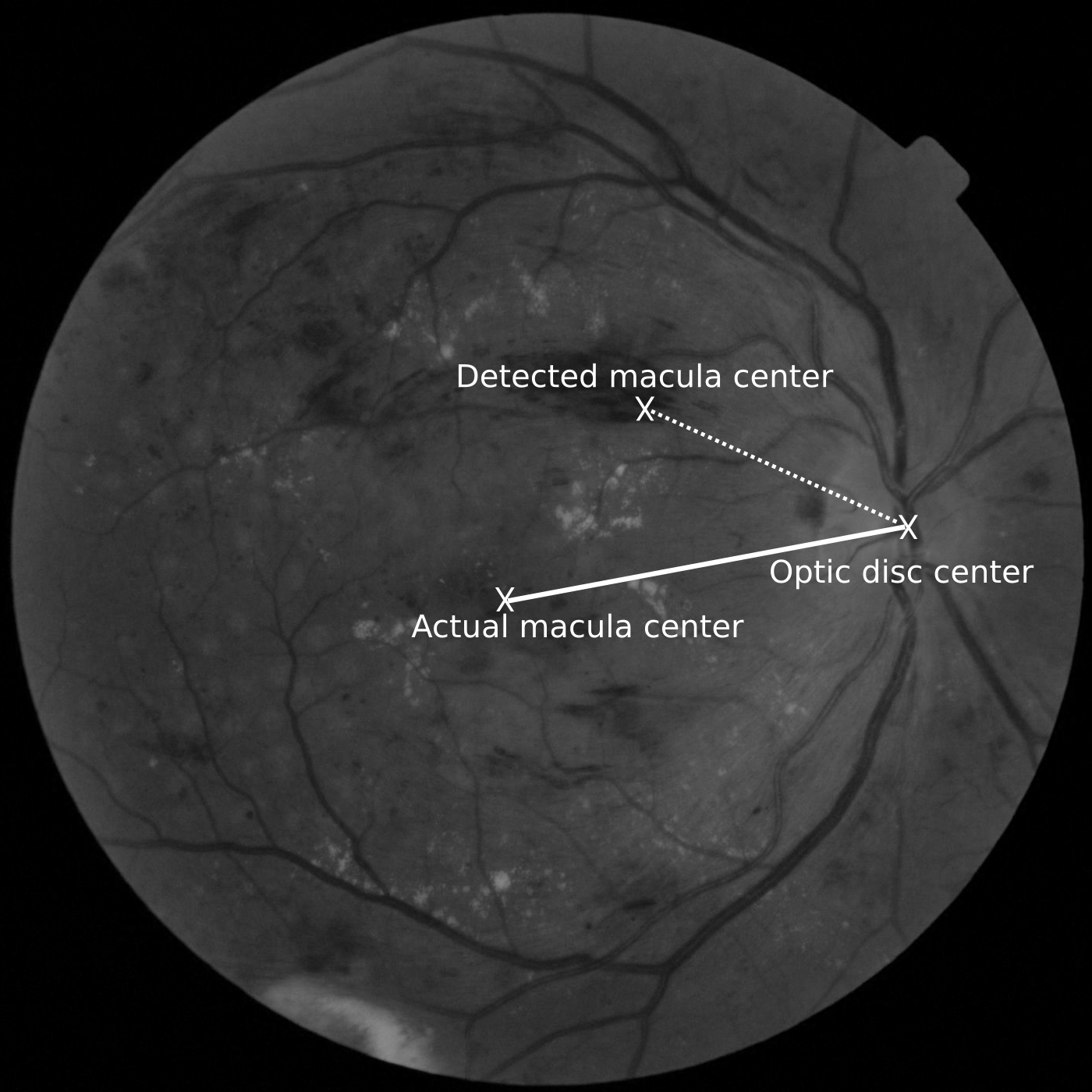}
	\caption{The difference between the actual and the detected optic disc and macula centers.}
	\label{fig:odcmc}
\end{figure}

$\bullet ~$ Since abnormalities can make it harder to detect certain anatomical landmarks in an image, $\chi_{17}$ represents the euclidean distance of the center of the macula and the center of the optic disc to provide important information regarding the patient's condition (see Figure \ref{fig:odcmc} for an example). 
This feature is also normalized with the diameter of the ROI.

$\bullet ~\chi_{18}$ is the result of the AM/FM-based classification, which is a non-negative scalar indicating the confidence of the detection of DR. The larger $\chi_{18}$, the higher the probability that DR is present.     

\begin{table}[!ht]
\caption{Features for DR grading.}
\centering
\begin{tabular}{|c|p{0.8\linewidth}|}
	\hline	
	Feature & Description of feature\\
	\hline
	$\chi_{0}$ & The result of quality assessment.\\
	\hline	
	$\chi_{1}$ & The result of pre-screening (non-severe DR / severe DR). \\
\hline
	$\chi_{2}-\chi_{7}$ & The number of MAs detected at confidence levels $\alpha = 0.5,\,\dots,\,1.0$, resp.\\
	\hline
	$\chi_{8} - \chi_{16}$ & The number of exudate pixels at confidence levels $\alpha = 0.1,\,\dots,\,1.0$, resp.\\
	\hline
	$\chi_{17}$ & The euclidean distance of the center of the macula and the center of the optic disc.\\
	\hline
	$\chi_{18}$ & The result of the AM/FM-based classification (No DR/DR).\\
	\hline
\end{tabular}

\label{tab:dr_features}
\end{table}

\subsection{Classifiers and energy functions}
\label{sec:classifiers}

We have considered the following classifiers as potential members of ensembles:

\begin{itemize}
	\item Alternating Decision Tree, 
	\item kNN,
	\item AdaBoost,
	\item Multilayer Perceptron,
	\item Naive Bayes,
	\item Random Forest, 
	\item SVM, 
	\item Pattern classifier \Citep{antal_icete2012}.
\end{itemize}


For ensemble selection, we have considered the following energy functions:
\begin{equation}
	\textit{Sensitivity} = \dfrac{TP}{TP+FN},
\label{eq:sen}
\end{equation}
\begin{equation}
	\textit{Accuracy} = \dfrac{TP+TN}{TP+FP+TN+FN},~\mbox{and}
\label{eq:acc}
\end{equation}
\begin{equation}
	\textit{F-score} = \dfrac{2TP}{2TP+FN+FP},
\label{eq:fscore}
\end{equation}
where $TP$, $FP$, $TN$, $FN$ represent the true and false positive and true and false negative classifications of the system, respectively. In the rest of the paper, when the functions (\ref{eq:sen}), (\ref{eq:acc}), (\ref{eq:fscore}) are set in italic, we refer to them as energy functions;
their normal typesetting forms mean the same function, but applied to evaluation purposes.

Note that to fit this realization to the general framework, the above features and classifiers should be considered as the ones 
$\chi_1, \dots, \chi_{18}$ and $D_1, \dots, D_8$ given in section \ref{sec:ens_learning_concepts}, respectively. Moreover, any of the energy functions (\ref{eq:sen}), (\ref{eq:acc}), (\ref{eq:fscore}) should be assigned to $E$ in section \ref{sec:selection}.

\subsection{Training and evaluation}

10-fold cross-validation have been used for both the training phase and for the evaluation of the ensembles. The figures given in section \ref{sec:exp_final} are the average values of the 10-fold cross-validation for the respective energy functions in each case on the Messidor database. 
To measure the performance of the ensembles, we disclose the following descriptive values: Sensitivity (\ref{eq:sen}), Accuracy (\ref{eq:acc}), and Specificity with the latter one is defined as:
\begin{equation}
	\mbox{Specificity} = \dfrac{TN}{TN+FP}.
\label{eq:spe}
\end{equation}
To compare our results with other approaches, we have fitted Receiver Operating Characteristic curves to the results and calculated the area under these curves ($AUC$) using JROCFIT \Citep{jrocfit}. 
We have evaluated the ensemble creation strategies in two scenarios:
\begin{itemize}
\item
\underline{R0 vs R1:} First, we have investigated whether the image contains early signs of retinopathy (R1) or not (R0), that is, 
$\Omega = \{\mbox{R0}, \mbox{R1} \}$ in Definition \ref{classifierdef}. Discriminating these two classes are the most challenging task of DR screening, since R1 usually contain only minor and visually less distinguishable signs of DR than advanced stages (R2, R3). \\
\item
\underline{No DR/DR:} Second, we have measured the classification performance of the ensembles between all diseased categories (R1, R2, R3) and the normal one (R0), that is, $\Omega = \{\mbox{R0}, \{\mbox{R1} \cup \mbox{R2} \cup \mbox{R3}\}\}$ in Definition \ref{classifierdef}.
\end{itemize}
 
\section{Results}
\label{sec:exp_final}

\subsection{Ensemble selection}

Tables \ref{tab:r0r1_forward} and \ref{tab:r0r1_backward} contain the Sensitivity, Specificity and Accuracy values corresponding to the different fusion strategies and search methods for the scenario R0 vs R1, while Tables \ref{tab:nodrdr_forward} and \ref{tab:nodrdr_backward} relate to the scenario No DR/DR, respectively. For both scenarios, the table entries corresponding to the most accurate ensembles are set in bold. For better comparison, we also disclose the accuracy values for the ensembles containing all classifiers in table \ref{tab:all}.

\begin{table}
\caption{DR grading results for scenario R0 vs R1 on the Messidor database with forward search method using different fusion strategies and energy functions. Each cell contains the Sensitivity/Specificity/accuracy of the best ensemble for the corresponding setup.}
\centering
\begin{tabular}{|p{0.15\linewidth}|c|c|c|}
\hline
\multicolumn{4}{|c|}{R0 vs R1 -- Forward search}\\
\hline
\diaghead(-3,2){\hskip3.2cm}{Fusion strategy}{Energy function}& \textit{Sensitivity} & \textit{Accuracy} & \textit{F-Score}\\ 
\hline
${\cal D}_{maj}$ & 98\%/82\%/83\% & 77\%/90\%/88\% & 75\%/89\%/86\%\\ 
\hline
${\cal D}_{wmaj}$ & 76\%/90\%/88\% & 83\%/88\%/87\% & 87\%/87\%/87\%\\
\hline 
${\cal D}_{avg}$ & 86\%/88\%/88\% & 77\%/88\%/86\% & 82\%/89\%/88\%\\ 
\hline
${\cal D}_{pro}$ & 74\%/90\%/86\% & 80\%/88\%/87\% & 79\%/89\%/87\%\\ 
\hline
${\cal D}_{min}$ & 74\%/90\%/87\% & 74\%/91\%/87\% & 85\%/88\%/88\%\\ 
\hline
${\cal D}_{max}$ &  77\%/90\%/87\% & 71\%/91\%/86\% & 81\%/88\%/87\%\\ 
\hline
\end{tabular}
\label{tab:r0r1_forward}
\end{table}

\begin{table}
\caption{DR grading results for scenario R0 vs R1 on the Messidor database with backward search method using different fusion strategies and energy functions. Each cell contains the Sensitivity/Specificity/accuracy of the best ensemble for the corresponding setup.}
\centering
\begin{tabular}{|p{0.15\linewidth}|c|c|c|}
\hline
\multicolumn{4}{|c|}{R0 vs R1 -- Backward search}\\
\hline
\diaghead(-3,2){\hskip3.2cm}{Fusion strategy}{Energy function} & \textit{Sensitivity} & \textit{Accuracy} & \textit{F-Score}\\ 
\hline
${\cal D}_{maj}$ & 88\%/87\%/87\% & 92\%/88\%/89\% & 84\%/89\%/88\%\\ 
\hline
${\cal D}_{wmaj}$ & 98\%/82\%/84\% & 85\%/88\%/88\%& 69\%/88\%/83\%\\
\hline 
${\cal D}_{avg}$ & 85\%/89\%/88\% & \textbf{94\%/90\%/90\%}& 93\%/90\%/90\%\\ 
\hline
${\cal D}_{pro}$ & 0\%/78\%/78\% & 0\%/79\%/80\% & 0\%/78\%/80\%\\ 
\hline
${\cal D}_{min}$ & 81\%/90\%/88\% & 83\%/89\%/88\% & 64\%/96\%/85\%\\ 
\hline
${\cal D}_{max}$ &  98\%/81\%/82\% & 98\%/81\%/83\% & 76\%/89\%/86\%\\ 
\hline
\end{tabular}
\label{tab:r0r1_backward}
\end{table}

\begin{table}
\caption{DR grading results for scenario No DR/DR on the Messidor database with forward search method using different fusion strategies and energy functions. Each cell contains the Sensitivity/Specificity/accuracy of the best ensemble for the corresponding setup.}
\centering
\begin{tabular}{|p{0.15\linewidth}|c|c|c|}
\hline
\multicolumn{4}{|c|}{No DR/DR -- Forward search}\\
\hline
\diaghead(-3,2){\hskip3.2cm}{Fusion strategy}{Energy function} & \textit{Sensitivity} & \textit{Accuracy} & \textit{F-Score}\\ 
\hline
${\cal D}_{maj}$ & 88\%/79\%/86\% & 91\%/76\%/88\% & 88\%/84\%/88\% \\ \hline 
${\cal D}_{wmaj}$ & 88\%/84\%/87\% & 88\%/88\%/87\% & 91\%/68\%/85\% \\ \hline 
${\cal D}_{avg}$ & 86\%/83\%/85\% & 88\%/85\%/88\% & 89\%/81\%/87\% \\ \hline 
${\cal D}_{pro}$ & 95\%/38\%/60\% & 85\%/83\%/85\% & 89\%/72\%/85\% \\ \hline 
${\cal D}_{min}$ & 80\%/95\%/80\% & 88\%/82\%/87\% & 87\%/78\%/86\% \\ \hline 
${\cal D}_{max}$ & 92\%/50\%/72\% & 90\%/76\%/87\% & 88\%/76\%/86\% \\ \hline 
\end{tabular}
\label{tab:nodrdr_forward}
\end{table}

\begin{table}[!htb]
\caption{DR grading results for scenario No DR/DR on the Messidor database with backward search method using different fusion strategies and energy functions. Each cell contains the Sensitivity/Specificity/accuracy of the best ensemble for the corresponding setup.}
\centering
\begin{tabular}{|p{0.15\linewidth}|c|c|c|}
\hline
\multicolumn{4}{|c|}{No DR/DR  -- Backward search}\\
\hline
\diaghead(-3,2){\hskip3.2cm}{Fusion strategy}{Energy function} & \textit{Sensitivity} & \textit{Accuracy} & \textit{F-Score}\\ 
\hline
${\cal D}_{maj}$ & 89\%/78\%/86\% & 90\%/80\%/89\% & 90\%/88\%/90\% \\ \hline 
${\cal D}_{wmaj}$ & 88\%/93\%/85\% & 86\%/83\%/85\% & 89\%/90\%/88\% \\ \hline 
${\cal D}_{avg}$& \textbf{90\%/91\%/90}\% & 87\%/80\%/86\% & 89\%/92\%/90\% \\ \hline 
${\cal D}_{pro}$ & 97\%/56\%/80\% & 88\%/85\%/88\% & 90\%/73\%/86\% \\ \hline 
${\cal D}_{min}$& 81\%/97\%/82\% & 81\%/97\%/82\% & 81\%/98\%/83\% \\ \hline 
${\cal D}_{max}$ & 93\%/68\%/86\% & 93\%/77\%/89\% & 89\%/83\%/88\% \\ \hline 
\end{tabular}
\label{tab:nodrdr_backward}
\end{table}

\begin{table}
\caption{DR grading results on the Messidor database with all of the classifiers included in the ensemble. Each cell contains the Sensitivity/Specificity/accuracy of the best ensemble for the corresponding setup.}
\centering
\begin{tabular}{|p{0.2\linewidth}|c|c|}
\hline
\multicolumn{3}{|c|}{All classifiers}\\
\hline
\diaghead(-3,2){\hskip3.2cm}{Fusion strategy}{Scenario} & R0 vs R1 & No DR/DR\\ 
\hline
${\cal D}_{maj}$ & 96\%/84\%/85\% & 88\%/79\%/86\%\\ 
\hline
${\cal D}_{wmaj}$ & 85\%/87\%/87\% & 88\%/84\%/87\%\\
\hline 
${\cal D}_{avg}$ & 80\%/88\%/87\% & 86\%/83\%/85\%\\ 
\hline
${\cal D}_{pro}$ & 100\%/78\%/78\% & 95\%/38\%/60\%\\ 
\hline
${\cal D}_{min}$ & 48\%/95\%/69\% & 80\%/95\%/80\%\\ 
\hline
${\cal D}_{max}$ & 95\%/79\%/80\% & 92\%/50\%/72\%\\ 
\hline
\end{tabular}
\label{tab:all}
\end{table}

Regarding the scenario R0 vs R1, from Table \ref{tab:r0r1_backward} we can see that the best performing ensemble achieved 94\% Sensitivity, 90\% Specificity and 90\% Accuracy using backward search, output fusion strategy ${\cal D}_{avg}$ and energy function \textit{Accuracy}. 
For the scenario No DR/DR, 90\% Sensitivity, 91\% Specificity and 90\% Accuracy are achieved with the same search method and fusion strategy (see Table \ref{tab:nodrdr_backward}). However, the energy function in this case is \textit{Sensitivity}. For a fair comparison, we also disclose the aggregated results for the energy functions and search methods in Tables \ref{tab:energy_r0r1}, and \ref{tab:search_r0r1} for the scenario R0 vs R1, and in Tables \ref{tab:energy_nodrdr}, and \ref{tab:search_nodrdr} for the scenario No DR/DR, respectively. 

\subsubsection{Energy functions}

For scenario R0 vs R1 we can state that while the energy functions \textit{Sensitivity} and \textit{Accuracy} have performed similarly,
\textit{F-score} has provided less accurate ensembles. For scenario No DR/DR all the three energy functions performed similarly. The difference in the effectiveness of the measure \textit{F-score} probably lies in the fact that the dataset for scenario R0 vs R1 is biased to R0, since it contains much more instances belonging to that class. That is, the energy functions \textit{Accuracy} and \textit{Sensitivity} look more robust for less balanced datasets. 

\begin{table}
\caption{Comparison of the energy functions for the scenario R0 vs R1.}
\centering
\begin{tabular}{|c|c|c|c|}
\hline
\multicolumn{4}{|c|}{R0 vs R1}\\
\hline
Energy function & Sensitivity & Specificity & Accuracy\\
 \hline
\textit{Sensitivity} &  86\% & 86\% & 86\%\\  
\hline
\textit{Accuracy}& 84\% & 88\% & 87\%\\ 
\hline  
\textit{F-score}  & 81\% & 88\% & 80\%\\ 
\hline  
\end{tabular}
\label{tab:energy_r0r1}
\end{table}

\begin{table}[!htb]
\caption{Comparison of the energy functions for the scenario No DR/DR.}
\centering
\begin{tabular}{|c|c|c|c|}
\hline
\multicolumn{4}{|c|}{No DR/DR}\\
\hline
Energy function & Sensitivity & Specificity & Accuracy\\
 \hline
\textit{Sensitivity} &  90\% & 79\% & 86\%\\  
\hline
\textit{Accuracy}& 88\% & 84\% & 87\%\\ 
\hline  
\textit{F-score}  & 88\% & 82\% & 87\%\\ 
\hline  
\end{tabular}
\label{tab:energy_nodrdr}
\end{table}

\subsubsection{Search methods}

As for the search methods, the accuracy of the forward and backward search method are similar. However, in both scenarios, the Sensitivity and Specificity values are more balanced for the backward strategy, which is desired for a grading system.

\begin{table}[!htb]
\caption{Comparison of the search methods for the scenario R0 vs R1.}
\centering
\begin{tabular}{|c|c|c|c|}
\hline
\multicolumn{4}{|c|}{R0 vs R1}\\
\hline
Search method & Sensitivity & Specificity & Accuracy\\
 \hline
Forward &  80\% & 89\% & 87\%\\  
\hline
Backward& 88\% & 86\% & 86\%\\ 
\hline  
All  & 84\% & 85\% & 81\%\\ 
\hline  
\end{tabular}
\label{tab:search_r0r1}
\end{table}

\begin{table}[!htb]
\caption{Comparison of the search methods for the scenario No DR/DR.}
\centering
\begin{tabular}{|c|c|c|c|}
\hline
\multicolumn{4}{|c|}{R0 vs R1}\\
\hline
Search method & Sensitivity & Specificity & Accuracy\\
 \hline
Forward &  90\% & 78\% & 87\%\\  
\hline
Backward& 88\% & 84\% & 87\%\\ 
\hline  
All  & 88\% & 71\% & 79\%\\ 
\hline  
\end{tabular}
\label{tab:search_nodrdr}
\end{table}

\subsubsection{Classifier output fusion strategies}

In Tables \ref{tab:fusion_r0r1}, and \ref{tab:fusion_nodrdr} the comparison of the fusion strategies can be observed. The experimental results indicate that ${\cal D}_{avg}$ is the most effective strategy for both scenarios. The aggregated results confirm this observation. However, 
${\cal D}_{maj}$ and ${\cal D}_{wmaj}$ have also provided similar results, suggesting possible alternative choices.  
 
\begin{table}
\caption{Comparison of classifier output fusion strategies for the scenario R0 vs R1.}
\centering
\begin{tabular}{|c|c|c|c|}
\hline
\multicolumn{4}{|c|}{R0 vs R1}\\
\hline
Fusion strategy & Sensitivity & Specificity & Accuracy\\ 
\hline
${\cal D}_{maj}$ & 87\% & 87\% &87\%\\ 
\hline
${\cal D}_{wmaj}$ &83\% & 87\% &86\%\\
\hline 
${\cal D}_{avg}$ & 85\% & 89\% &88\%\\ 
\hline
${\cal D}_{pro}$ & 33\% & 82\% &82\%\\ 
\hline
${\cal D}_{min}$ & 73\% & 92\% &85\%\\ 
\hline
${\cal D}_{max}$ & 85\% & 86\% &84\%\\ 
\hline
\end{tabular}
\label{tab:fusion_r0r1}
\end{table}

\begin{table}
\caption{Comparison of classifier output fusion strategies for the scenario No DR/DR.}
\centering
\begin{tabular}{|c|c|c|c|}
\hline
\multicolumn{4}{|c|}{No DR/DR}\\
\hline
Fusion strategy & Sensitivity & Specificity & Accuracy\\ 
\hline
${\cal D}_{maj}$ & 89\% & 80\%& 88\%\\ 
\hline
${\cal D}_{wmaj}$ &88\% & 83\%& 87\%\\
\hline 
${\cal D}_{avg}$ & 88\% & 84\%& 88\%\\ 
\hline
${\cal D}_{pro}$ & 91\% & 69\%& 81\%\\ 
\hline
${\cal D}_{min}$ & 84\% & 89\% &84\%\\ 
\hline
${\cal D}_{max}$ & 91\% & 77\% &85\%\\ 
\hline
\end{tabular}
\label{tab:fusion_nodrdr}
\end{table}

To conclude on the analysis of ensemble selection approaches, it can be stated that backward ensemble search method with energy functions 
\textit{Sensitivity} or \textit{Accuracy} and fusion strategy ${\cal D}_{avg}$ can be recommended for ensemble selection for automatic DR screening.

\subsection{Comparison with other automatic DR screening systems}

It is challenging to compare our approach with other methods. As we can see in Table \ref{tab:compare_systems}, most research groups not only evaluated their approach on private datasets, but the proportion of images showing signs of DR is also completely different. Moreover, the most meaningful measure, the area under the ROC curves is not always disclosed either. However, the proposed approach has provided significantly better performance then the other state-of-the-art techniques regarding the clinically important measures. Also note that this comparison was able to be made
only for the scenario No DR/DR because of the lack of data for scenario R0 vs R1 from the other systems.

\begin{table}[!htb]
\caption{Comparison of automatic DR screening systems.}
\centering
\begin{tabular}{|c|c|c|c|c|}
\hline
System & Cases having DR & Sensitivity & Specificity & AUC\\
\hline
\Citep{system} & 4.8\% & 84\% & 64\% & 0.84\\
\hline
\Citep{early} & 4.96\% &  N/A &  N/A& 0.86\\ 
\hline
\Citep{Jelinek2006} & 30\% & 85\% & 90\% &  N/A\\ 
\hline
\Citep{antal_tbme2012} & 46\% & 76\% & 88\% & 0.90\\ 
\hline
\Citep{Philip} & 37.5\% & 90.5\% & 54.7\% & N/A\\
\hline
\Citep{largescale} & 35.88\% & 87\% & 50.4\% &  N/A\\ 
\hline
\Citep{Agurto} & 74.43\% &  N/A &  N/A & 0.81\\ 
\hline
\Citep{Agurto} & 76.26\% &  N/A &  N/A & 0.89\\ 
\hline
Proposed & 46\% & 90\% & 91\% & 0.989\\ 
\hline  
\end{tabular}
\label{tab:compare_systems}
\end{table}

In \Citep{antal_tbme2012}, we have reported grading results for the dataset Messidor based on only MA detection
for both scenarios. The comparative results between the proposed system and \Citep{antal_tbme2012} are given in Table \ref{tab:compare_messidor_r0r1} for the scenario R0 vs R1, and in Table \ref{tab:compare_messidor_nodrdr} for the 
scenario No DR/DR, respectively. To highlight the efficiency of the ensemble-based approach, we have 
included the results corresponding to a single classifier based decision, as well. As we can see, the proposed 
system outpeforms both \Citep{antal_tbme2012} and the single classifier approach. 
It is also interesting to note that the single classifier approach clearly performs better than \Citep{antal_tbme2012}, 
which is based solely on the detection of MAs. This observation also confirms the necessity of the wide range of components. 

\begin{figure}[htb]
	\centering
		\includegraphics[width=\linewidth]{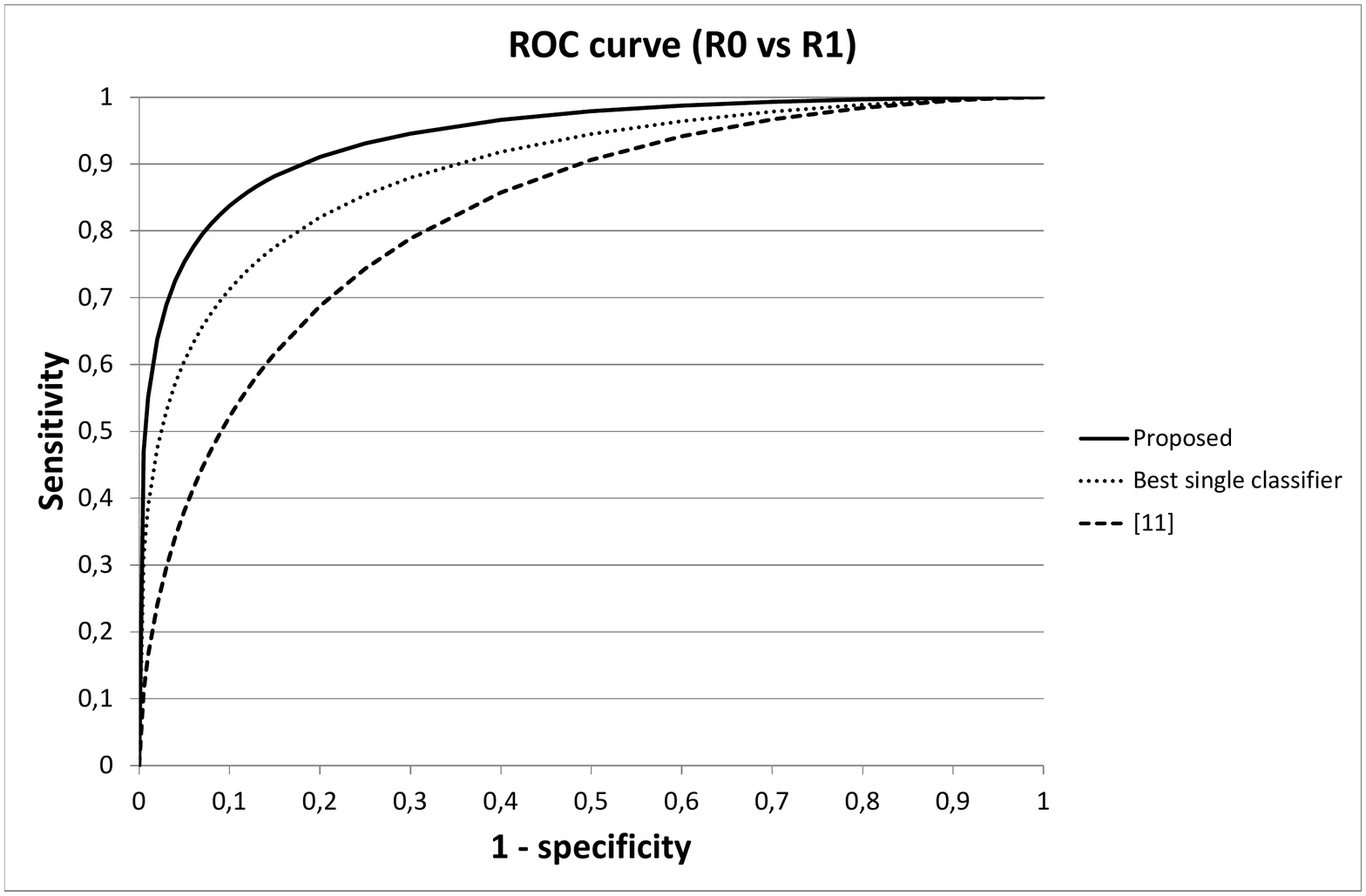}
	\caption{ROC curves of automatic DR screening systems evaluated on the Messidor dataset for the scenario R0 vs R1.}
	\label{fig:messidor_roc_r0r1}
\end{figure}

\begin{table}[!htb]
\caption{Comparison of automatic DR screening systems evaluated on the Messidor dataset for the scenario R0 vs R1.}
\centering
\begin{tabular}{|c|c|c|c|c|}
\hline
\multicolumn{5}{|c|}{R0 vs R1}\\
\hline
System &  Sensitivity & Specificity & Accuracy & AUC\\
 \hline
\Citep{antal_tbme2012} &  97\% & 14\% & 32\% & 0.826\\  
\hline
Best single classifier& 85\% & 87\% & 86\% & 0.893\\ 
\hline  
Proposed & 94\% & 90\% & 90\% & 0.942\\ 
\hline  
\end{tabular}
\label{tab:compare_messidor_r0r1}
\end{table}

\begin{figure}[htb]
	\centering
		\includegraphics[width=\linewidth]{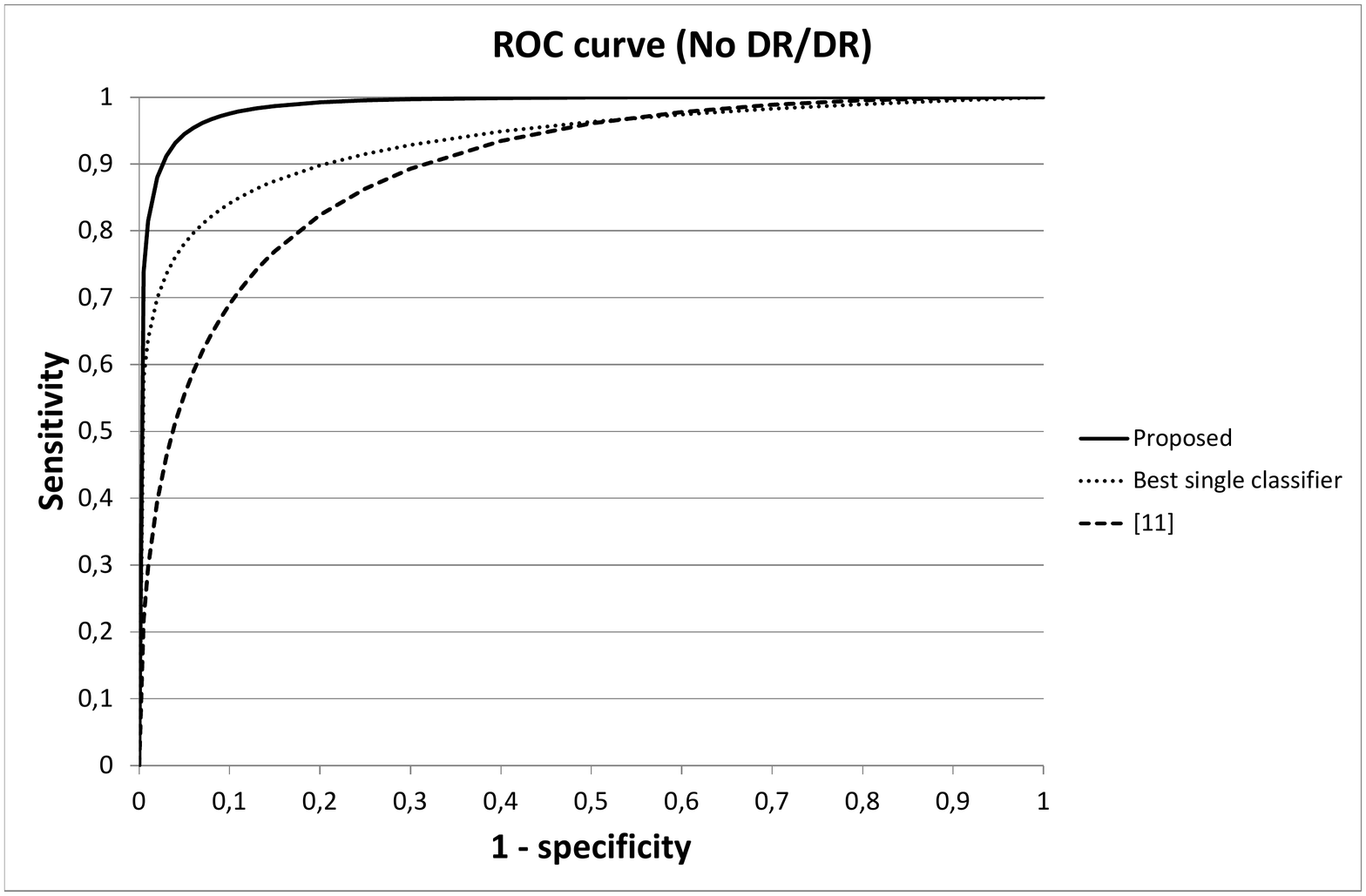}
	\caption{ROC curves of automatic DR screening systems evaluated on the Messidor dataset for the scenario No DR/DR.}
	\label{fig:messidor_roc_nodrdr}
\end{figure}

\begin{table}[!htb]
\caption{Comparison of automatic DR screening systems evaluated on the Messidor dataset for the scenario No DR/DR.}
\centering
\begin{tabular}{|c|c|c|c|c|}
\hline
\multicolumn{5}{|c|}{No DR/DR}\\
\hline
System &  Sensitivity & Specificity & Accuracy & AUC\\
 \hline
\Citep{antal_tbme2012} &  76\% & 88\% & 82\% & 0.90\\  
\hline
Best single classifier& 90\% & 81\% & 86\% & 0.936\\ 
\hline  
Proposed & 90\% & 91\% & 90\% & 0.989\\ 
\hline  
\end{tabular}
\label{tab:compare_messidor_nodrdr}
\end{table}

\section{Conclusion} 
\label{sec:conclusion} 

In this paper, we have proposed an ensemble-based automatic DR screening system. Opposite to the state-of-the-art methods, we have used image-level, lesion-specific and anatomical components at the same time. To strengthen the reliability of our approach, we have created an ensemble of classifiers. We have discussed extensively on how an efficient ensemble for such a task can be found. Our approach has been validated on the publicly available dataset Messidor, where an outstanding 0.989 area under the ROC curve is achieved. The presented results outperform the current state-of-the-art techniques, which can be reasoned by the well-known observation that ensemble-based systems often lead to higher accuracies. It is also worth noting that our system can be very easily extended by adding more/other components and classifiers. The sensitivity/specificity results (90\%/91\%) we have achieved are also close to the recommendations of the British Diabetic Association (BDA) (80\%/95\%) for DR screening \Citep{bda}.
 
\section*{Acknowledgment}
This work was supported in part by the project TAMOP-4.2.2.C-11/1/KONV-
2012-0001 supported by the European Union, co-financed by the European Social Fund; the OTKA grant NK101680; and by the TECH08-2 project DRSCREEN - Developing a computer based image processing system for diabetic retinopathy screening of the National Office for Research and Technology of Hungary (contract no.: OM-00194/2008, OM-00195/2008, OM-00196/2008) and by the European Union and the State of Hungary, co-financed by the European Social Fund in the framework of TÁMOP-4.2.4.A/ 2-11/1-2012-0001 ‘National Excellence Program’.

\bibliographystyle{elsarticle-harv}
\bibliography{refs_rev}







\end{document}